\documentclass{article}

\usepackage{microtype}
\usepackage{graphicx}
\usepackage{subfigure}
\usepackage{booktabs} 
\usepackage[inline]{enumitem}   
\usepackage{subcaption}
\usepackage{framed} 

\usepackage{hyperref}

\usepackage[accepted]{icml2024}

\usepackage{amsmath}
\usepackage{amssymb}
\usepackage{mathtools}
\usepackage{amsthm}

\usepackage[capitalize,noabbrev]{cleveref}

\theoremstyle{plain}

\theoremstyle{definition}

\theoremstyle{remark}

\usepackage[textsize=tiny]{todonotes}

\icmltitlerunning{Improving Rule-based Reasoning in LLMs using Neurosymbolic Representations}

\begin{document}

\twocolumn[
\icmltitle{Improving Rule-based Reasoning in LLMs using Neurosymbolic Representations}

\begin{icmlauthorlist}
    \icmlauthor{Varun Dhanraj}{uwcs,ctn}
    \icmlauthor{Chris Eliasmith}{ctn}
\end{icmlauthorlist}

\icmlaffiliation{uwcs}{School of Computer Science, University of Waterloo, Waterloo, Canada}
\icmlaffiliation{ctn}{Centre for Theoretical Neuroscience, University of Waterloo}

\icmlcorrespondingauthor{Varun Dhanraj}{vdhanraj@uwaterloo.ca}

\vskip 0.3in
]

\printAffiliationsAndNotice{}

\begin{abstract}
Large language models (LLMs) continue to face challenges in reliably solving reasoning tasks, particularly tasks that involve precise rule following, as often found in mathematical reasoning tasks. This paper introduces a novel neurosymbolic method that improves LLM reasoning by encoding hidden states into neurosymbolic vectors, enabling problem-solving within a neurosymbolic vector space. The results are decoded and merged with the original hidden state, significantly boosting the model’s performance on numerical reasoning tasks. By offloading computation through neurosymbolic representations, this method enhances efficiency, reliability, and interpretability. Our experimental results demonstrate an average of 88.6\% lower cross-entropy loss and 15.4 times more problems correctly solved on a suite of mathematical reasoning tasks compared to chain-of-thought prompting and supervised fine-tuning (LoRA), while not hindering the LLM’s performance on other tasks. We make our code available at \textbf{Neurosymbolic LLM}\footnote{\url{https://github.com/vdhanraj/Neurosymbolic-LLM}}

\end{abstract}

\begin{figure}[ht]
\vskip 0.2in
\begin{center}
\centerline{\includegraphics[width=\columnwidth]{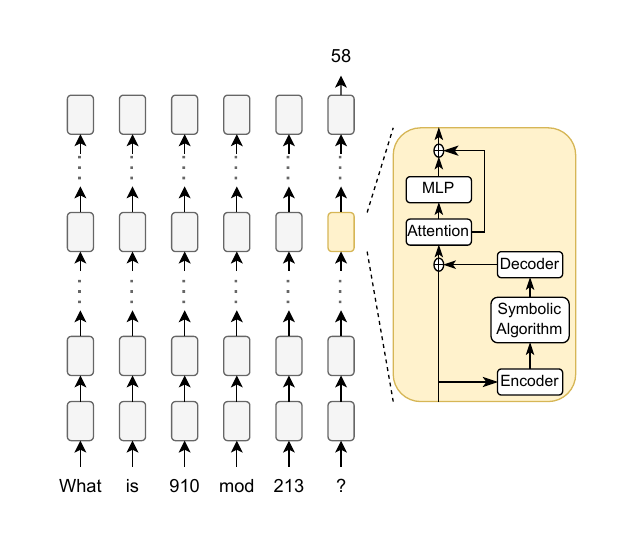}}
\caption{A diagram of our method, showing how LLM hidden states are converted into compositional neurosymbolic representations. The encoder network converts the LLM hidden state to a neurosymbolic vector which can be queried to obtain the ones, tens, and hundreds digit of each number, as well as the type of problem being asked. This information is used by the neurosymbolic algorithm to find a solution to the problem, which the decoder converts from a neurosymbolic vector into an LLM hidden state vector, which is then added to the original LLM hidden state.}
\label{system_diagram}
\end{center}
\vskip -0.2in
\end{figure}

\section{Introduction} \label{sec:intro}
Despite the remarkable progress in deep learning, significant gaps remain between the strengths of deep learning-based models and traditional symbolic reasoning systems \cite{gsmsymbolic, symbolicReasoning}. Deep learning excels at intuition and pattern recognition, leveraging large datasets to make flexible, context-aware predictions. However, these models often suffer from issues such as hallucinations and a lack of reliability, especially when solving tasks that require strict rule-following and logical consistency \citep{lin2023false, chen2023logical}. In contrast, symbolic reasoning methods provide precision and reliability, but they struggle to scale to complex and noisy real-world problems.

This dichotomy has fueled a growing interest in merging the strengths of these two paradigms. Many integrated approaches aim to leverage the intuition and adaptability of large language models (LLMs) while incorporating the rigor and interpretability of symbolic reasoning \citep{logiclm2023,linc2023,chatlogic2024,symbolicai2024}. For example, approaches such as deep learning-guided program synthesis aim to use LLMs to generate complex algorithms by producing code for various candidate programs that could solve abstract reasoning problems \cite{arc}. While this approach demonstrates the potential of combining neural network-based pattern recognition with symbolic algorithms for programmatic reasoning, it remains constrained to token-level operations and fails to leverage the richer and more complex information embedded within the LLM’s hidden states.

In this paper, we introduce a novel method that extends the capabilities of LLMs by encoding their hidden states into structured symbolic vector representations. Unlike previous work focusing on token-level program synthesis, our approach directly integrates symbolic algorithms within the neural model by running them in a symbolic space derived from the LLM's internal representations. This innovation bridges the gap between neural and symbolic reasoning by extracting inputs from the LLM's hidden state and operating directly on a structured, interpretable representation of the problem.

Our contributions include:

\begin{itemize}
\item \textbf{A Neurosymbolic Method for LLMs}: This work represents a first step toward integrating symbolic reasoning into LLMs 
. We explore the ability of symbolic algorithms to operate within a symbolic space constructed from the LLM’s latent representations.
\item \textbf{Symbolic Representations from Hidden States}: We demonstrate the feasibility of decoding state information from LLM hidden layers into structured, compositional symbolic representations using Vector Symbolic Algebras (VSAs). These representations enable rule-based manipulation of mathematical and logical constructs.
\item \textbf{Improved Performance on Rule-Based Tasks}: By leveraging neurosymbolic processing, our approach achieves significant improvements in accuracy and interpretability on numerical reasoning tasks, outperforming methods like chain-of-thought (CoT) prompting and Low-Rank Adaptation (LoRA) fine-tuning.
\end{itemize}

This work enables symbolic algorithms to run directly within neural networks, laying the groundwork for more advanced neurosymbolic systems that balance the adaptability of LLMs with the reliability of symbolic reasoning. By integrating neurosymbolic algorithms and decoding hidden state information into structured neurosymbolic representations, we aim to unlock new possibilities for solving complex, rule-based problems previously only solvable via symbolic approaches such as program synthesis.

\section{Related Work} \label{sec:related_work}

\subsection{Linear Probes} \label{subsec:linear_probes}
Linear probes are widely used tools for interpreting the internal representations of LLMs \cite{hewitt2019structural, liu2019linguistic}. They involve training a lightweight, linear mapping from a model's hidden states to specific properties of interest, such as linguistic features or numerical values. By analyzing how well these linear mappings perform, researchers can infer what information is encoded in the model's hidden states. For numerical reasoning, linear probes have been used to represent values by extracting information directly from hidden states \cite{elhage2021mathematical}. 

Previous work has extended this approach with digit-specific circular probes, which attempt to decompose numerical representations into their constituent digits using circular algebra \cite{elhage2022solu}. However, such methods generally exhibit lower accuracy compared to traditional linear probes and are limited in scope. Specifically, circular probes can only detect numbers and lack the ability to discern operations or broader semantic relationships.

In contrast, the method proposed in this work addresses these limitations by leveraging vector symbolic algebras (VSAs) to encode both numbers and operations. VSA-based representations offer dynamic scalability, allowing new functionality to be integrated without retraining the probe. Our approach is thus particularly well-suited for complex numerical reasoning tasks that require flexible and interpretable encodings.

\subsection{Sparse Autoencoders} \label{subsec:sae}
Sparse autoencoders (SAEs) are a class of unsupervised learning methods designed to parse high-dimensional data, such as the hidden states (also called activations) of LLMs, into sparse, monosemantic components \cite{olah2020zoom, le2021probing}. These components, often referred to as ``concepts,''
 are linearly combined to reconstruct the original input data \cite{elhage2021mathematical}. SAEs have been used to identify which latent features in an LLM are active during specific tasks, enabling researchers to explore the internal representations of the model. Furthermore, SAEs can be used to steer LLMs by selectively amplifying or suppressing certain concepts, providing a powerful tool for interpretability and control.

Despite these advantages, SAEs face notable limitations. First, the concepts learned by SAEs are not guaranteed to be atomic or aligned with structured representations, such as individual digits in numerical data. This ambiguity makes SAEs less suitable for tasks that require precise decomposition of hidden states. Second, the representations learned by SAEs are probabilistic and emergent, determined during training without external constraints, which complicates their use in symbolic algorithms \cite{olah2020zoom, elhage2021mathematical}. 

Additionally, the concepts extracted by SAEs are typically non-interpretable by default, requiring manual inspection of activations to identify their semantic meaning \cite{olah2020zoom, elhage2021mathematical}. While this can provide insights into LLM internals, it is labor-intensive and less systematic than the interpretable symbolic representations proposed in this paper. Finally, SAEs operate in an unsupervised setting, whereas the approach presented here uses supervised learning to enforce specific properties on the learned representations. This trade-off introduces inductive biases but ensures that the resulting encodings are structured and interpretable, facilitating their use in numerical reasoning tasks.

\section{Vector Symbolic Algebras} \label{sec:vsa}

Vector Symbolic Algebras (VSAs) are a family of algebras for constructing compositional symbol-like representations within a fixed-dimensional vector space. In this work, we use the HRR VSA \cite{plate1995holographic} to interpret the internal representations of LLMs and encode numerical reasoning tasks. VSAs enable the creation of neurosymbolic vectors that represent both data and operations, facilitating compact, interpretable, and algebraically manipulable representations.

VSAs are characterized by three key operations: \textit{bundling}, \textit{binding}, and \textit{similarity}, which allow for the creation and comparison of compositional representations:
\begin{itemize}
    \item \textbf{Bundling:} Combines multiple vectors to represent a set of elements (implemented as vector addition in HRRs).
    \item \textbf{Binding:} Represents associations between elements (implemented as circular convolution in HRRs).
    \item \textbf{Similarity:} Compares two vectors to determine how closely they match (implemented as the dot product in HRRs).
\end{itemize}

The binding operation, circular convolution, is formally defined as:
\begin{equation}
   (\mathbf{x} \circledast \mathbf{y})_i := \sum_{j=1}^{d} x_j y_{((i-j) \bmod d)+1}, \; i \in \{1, 2, \ldots, d\}.
   \label{eq:convolution}
\end{equation}

\subsection{Encoding Compositional Data} \label{subsec:comp_data}

VSAs allow compositional data to be encoded in a fixed-dimensional vector. For example, to represent a three-digit number, we assign vectors to the digits (\(0\) through \(9\)) and their respective place values (\textit{ones}, \textit{tens}, \textit{hundreds}). Using randomly initialized vectors, we can represent the number \(842\) as:
\begin{align}
\mathbf{x} = \mathbf{hundreds} \circledast \mathbf{8} + \mathbf{tens} \circledast \mathbf{4} + \mathbf{ones} \circledast \mathbf{2}.
\label{eq:SP_num_rep}
\end{align}

This process generalizes to encode multiple numbers and their relationships. For instance, we encode the query ``What is \(842 \text{ mod } 910\)?'' as:
\begin{align}
\mathbf{x} = &\; \mathbf{n_1} \circledast (\mathbf{hundreds} \circledast \mathbf{8} + \mathbf{tens} \circledast \mathbf{4} + \mathbf{ones} \circledast \mathbf{2}) \notag \\
&+ \mathbf{n_2} \circledast (\mathbf{hundreds} \circledast \mathbf{9} + \mathbf{tens} \circledast \mathbf{1} + \mathbf{ones} \circledast \mathbf{0}) \notag \\
&+ \mathbf{problem\_type} \circledast \mathbf{modulo},
\label{eq:SP_prob_rep}
\end{align}

To incorporate the structure of numbers (i.e., their ordering relations), digits can be encoded systematically \cite{choo2010a, Eliasmith2013}. For instance, a digit can be constructed by binding the vector for \(1\) with itself multiple times, e.g., \(\mathbf{3} = \mathbf{1} \circledast \mathbf{1} \circledast \mathbf{1}\). This scales well to higher values as long as we impose the further restriction that the base vector (e.g., 1) is unitary (i.e., all frequency components have a magnitude of 1). Similarly, we construct place values like \textit{tens} and \textit{hundreds} as repeated bindings of \textit{ones}, e.g., \(\mathbf{tens} = \mathbf{ones} \circledast \mathbf{ones}\). This systematic approach ensures desired numerical relations exist between the neurosymbolic vectors.

\subsection{Unbinding and the Pseudo-Inverse} \label{subsec:unbinding}
VSAs support \textit{unbinding}, which allows specific components of a compositional representation to be queried. For HRRs, unbinding is achieved by binding with the inverse of a neurosymbolic vector. Specifically, we use the pseudo-inverse of a vector \(\mathbf{y}\), denoted \(\mathbf{y}^{\dagger}\), which is obtained by flipping the order of all but the first element:
\begin{align}
\mathbf{y}^\dagger = (y_1, y_d, y_{d-1}, \dots, y_2),
\end{align}
where \(d\) is the dimensionality of the vector.

If \(\mathbf{z} = \mathbf{x} \circledast \mathbf{y}\), then unbinding \(\mathbf{z}\) with \(\mathbf{y}^{\dagger}\) approximately retrieves \(\mathbf{x}\):
\begin{align}
\mathbf{x} \approx \mathbf{y}^\dagger \circledast \mathbf{z}.
\end{align}

The unbinding operation can be used to extract specific components from a neurosymbolic representation. For example, consider the neurosymbolic vector described in (\ref{eq:SP_prob_rep}). If we want to query the hundreds digit of the second number (\(910\)), we unbind \(\mathbf{x}\) with \(\mathbf{n_2}\) and then with \(\mathbf{hundreds}\):
\begin{align}
\mathbf{result} = \mathbf{hundreds}^{\dagger} \circledast (\mathbf{n_2}^{\dagger} \circledast \mathbf{x}).
\end{align}
The resulting vector \(\mathbf{result}\) will have maximum similarity with \(\mathbf{9}\), corresponding to the hundreds digit of the second number.

\subsection{Vector Orthogonality and Capacity} \label{subsec:capacity}
One strength of a VSA-based approach is the ability to work with a large number of roughly orthogonal vectors, which facilitates the construction of complex structured representations. For a \(d\)-dimensional vector space, the number of vectors that maintain a similarity below a threshold \(\epsilon\) scales as:
\begin{align}
N \propto \exp\left(\alpha d \epsilon^2\right),
\end{align}
where \(\alpha\) is a constant derived from spherical code packing and the Kabatiansky–Levenshtein bound \cite{kabatiansky1978bounds, plate1995holographic}. This relationship is valid when \(\epsilon \sim \mathcal{O}(1 / \sqrt{d})\), and in this regime, the capacity grows exponentially with \(d\), enabling the representation of highly complex compositional structures.

By combining the properties described in this section, VSAs provide a robust framework for encoding and manipulating numerical reasoning representations, offering scalability, compositionality, and interpretability.

\section{Methodology} \label{sec:methodology}

Our method consists of three stages, which together provide an approach for enhancing the reasoning capabilities of LLMs through neurosymbolic processing. These stages are:
\begin{enumerate}
    \item Prompting the LLM with mathematical reasoning problems and gathering the hidden states from the model's layers.
    \item Encoding the gathered hidden states into neurosymbolic VSA representations that capture key features of the reasoning process.
    \item Applying rule-based algorithms to the representations, then decoding the results back into the LLM to generate final solutions.
\end{enumerate}

Next, we describe the dataset used in this study, before returning to describe each of these stages in more detail.

\subsection{Dataset} \label{subsec:dataset}

We release a \emph{formally specified, procedurally generated} benchmark,
the \textbf{Symbolic‑Math Dataset}\footnote{\url{https://github.com/vdhanraj/Symbolic-Math-Dataset}},
to foster reproducible evaluation of arithmetic reasoning in LLMs.
The dataset is open-source (MIT license) and fully regenerable, enabling reproducibility and scaling to more complex queries of the same arithmetic form (i.e., operations over arbitrarily many digits).

\paragraph{Construction.}
In this study, each example is built by
\emph{(i)} sampling two independent three‑digit integers
($x,y\in\{0,\dots,999\}$) and  
\emph{(ii)} sampling a problem type $t$ from a fixed set of $p=10$ symbolic
operations (listed below).
To ensure every operand and result remains a single sub‑word token in
LLaMA‑3, we mod‑reduce any outcome that exceeds three decimal digits:
e.g.\ $(932\!\times\!152)\bmod 1000 = 816$.
The instance is rendered as a natural‑language question such as

\begin{center}
\small
\texttt{``What is 932 times 152 mod 1000?''}
\end{center}

and paired with the numeric answer encoded as a single token. The problem types used in this study are:
\begin{enumerate}[label=(\arabic*), itemsep=0.4ex, leftmargin=*]
    \item \textbf{Modulo:} $x \bmod y$,
    \item \textbf{Multiplication:} $(x\cdot y) \bmod 10^{3}$,
    \item \textbf{GCD:} $\gcd(x,y)$,
    \item \textbf{LCM:} $\operatorname{lcm}(x,y)\bmod 10^{3}$,
    \item \textbf{Square Modulo:} $x^{2}\bmod y$,
    \item \textbf{Bitwise AND:} $\mathrm{int}(\mathrm{bin}(x)\,\&\,\mathrm{bin}(y))$,
    \item \textbf{Bitwise XOR:} $\mathrm{int}(\mathrm{bin}(x)\,\oplus\,\mathrm{bin}(y))$,
    \item \textbf{Bitwise OR:}  $\mathrm{int}(\mathrm{bin}(x)\,\lor\,\mathrm{bin}(y))$,
    \item \textbf{Addition:} $x + y$,
    \item \textbf{Integer Division:} $x // y$.
\end{enumerate}

Separate training, validation, and test splits are procedurally generated. The training and validation sets exclude \emph{addition} and \emph{integer division}, which are included only in the test set to evaluate out‑of‑distribution generalization.

\paragraph{Prompting format.}
In our study, each test query is presented in a few-shot format with two in-context exemplars of the same problem type, preceding the target question.
This consistent demonstration style encourages the model to learn the syntactic and arithmetic patterns of the task from examples alone, promoting the model to provide responses in a consistent and easy to evaluate format.



\subsection{Prompting and Gathering Hidden States} \label{subsec:gather_states}

In the first stage of our method, the LLM is presented with mathematical reasoning problems formulated as natural language questions. For each prompt, we extract the hidden state of the most recent token from a designated layer of the LLM, capturing an intermediate representation of the reasoning process.

For this study, we use LLaMA 3.1 8B, which features 4096-dimensional hidden state vectors at each of its 32 layers. Each layer consists of a self-attention mechanism, a feed-forward MLP, skip connections, and RMS normalization \cite{llama3}. Our approach records the hidden states just before they are processed by the selected layer, preserving an unaltered view of the model's internal representations at that stage.

\subsection{Encoding Hidden States} \label{subsec:encoding}

The second stage, after prompting, involves converting the hidden states of the LLM into neurosymbolic vector representations. For this purpose, we train a linear encoder network designed to map the hidden states recorded during the forward pass into neurosymbolic vectors that represent the problem's key components: the two input numbers and the operation type (see Figure~\ref{system_diagram}). For problems involving mod \(1000\) to truncate the final three digits, the \(1000\) is not represented as an input number, but instead is tied to a problem type (e.g., multiplication problem types will always apply modulo \(1000\) to the final answer). The symbolic vectors are structured using the framework described in Section~\ref{subsec:comp_data}. The encoder is trained using a root mean squared error (RMSE) loss, with the objective of minimizing the difference between the predicted and true symbolic vectors.

\subsection{Decoding Neurosymbolic States} \label{subsec:decoding}

Once the encoder network is trained, a corresponding linear decoder network is trained to reverse this mapping. The decoder network takes symbolic vectors as input, reconstructs the LLM's hidden state, and is optimized to minimize the RMSE loss between the original and reconstructed hidden states. The input dataset for the decoder training is generated by converting hidden states from the LLM into symbolic vectors using the trained (and now frozen) encoder network.

After training, both the encoder and decoder networks are included in the LLM (as shown in Figure \ref{system_diagram}) to assist in solving mathematical reasoning problems. The inference process begins by encoding the hidden state of the designated LLM layer into a neurosymbolic vector. This vector is then queried to determine the problem type, which dictates the selection of an appropriate rule-based Python function. If the queried problem type is not sufficiently similar to any the problem types encountered during training, the decoder is bypassed, and the LLM proceeds with its standard forward pass. Otherwise, the predefined rule-based function is applied to the extracted input values from the neurosymbolic vector, generating a new neurosymbolic representation containing the computed solution. This solution vector is then decoded back into an LLM-compatible hidden state via the decoder network, allowing the model to incorporate the computed result into its forward pass.

The output of the decoder is linearly combined with the original hidden state at the intervention layer to form the final hidden state. This linear mixing is performed using a 50-50 ratio, such that the resulting hidden state is:
\[
h_{\text{final}} = 0.5 \cdot h_{\text{decoder}} + 0.5 \cdot h_{\text{original}},
\]
where \(h_{\text{decoder}}\) is the output of the decoder network and \(h_{\text{original}}\) is the LLM's hidden state at the same layer.

Note that the layer at which the encoder generates the neurosymbolic vector from the hidden state does not need to be the same layer at which the decoder network uses the solution neurosymbolic vector to impact the hidden state of the LLM. In fact, multiple decoder layers may be trained and used to influence the hidden state of the LLM at different layers using the solution symbolic vector. For simplicity, we only choose layer 17's encoder and decoder network to both generate the neurosymbolic vector of the problem and to apply intervention to the forward pass of the LLM. The reasoning in choosing layer 17 is discussed further in Section~\ref{subsec:encoder_perf}.

Although the decoder networks are pretrained to reconstruct hidden states corresponding to symbolic vectors, their direct use during the LLM's forward pass may disrupt the algorithm being executed by the LLM, leading to degraded performance. This disruption occurs because the pretrained decoder networks map neurosymbolic vectors containing problem solutions directly into the LLM’s hidden states. However, the LLM’s original forward pass has hidden states that encode the problem inputs rather than the solution. Replacing the hidden states with representations of the solution can interfere with subsequent layers of the LLM, which expect input representations to align with the problem's original structure.

To address this issue, the decoder networks are fine-tuned by calculating the cross entropy loss of the logits of the correct token during the LLM's forward pass. This loss measures the discrepancy between the model's predicted output and the expected solution, allowing the decoder networks to adapt their mappings. The fine-tuning process ensures that the modified hidden states generated by the decoder networks not only represent the solution but also align with the LLM's internal expectations, enabling the model to generate correct outputs.

Fine-tuning the decoder layers achieves two objectives:
\begin{enumerate}[label=(\arabic*), itemsep=0.5ex, parsep=0pt]
    \item It teaches the decoder networks to map solution neurosymbolic vectors into hidden states that align with the LLM's forward-pass expectations.
    \item It mitigates disruptions to the LLM's computations caused by direct interventions in hidden states, ensuring the model generates correct outputs.
\end{enumerate}

Without fine-tuning, decoder outputs may cause the model to deviate from its learned reasoning pathways, leading to errors. By fine-tuning, the decoder networks adapt to the model’s computational context, improving overall performance in mathematical reasoning tasks.

\subsection{Computational complexity.}
Although our method introduces additional neurosymbolic processing steps, its computational overhead is minimal.
As detailed in Appendix~\ref{sec:complexity}, the total runtime cost per forward pass through the neurosymbolic block is
\[
\Theta\!\bigl(d\,v + v\log v\bigr),
\]
which is independent of the sequence length~$n$ and number of layers~$L$, and is asymptotically dominated by the standard key--value cached transformer cost of $\mathcal{O}(L(n\,d + d^2))$.
The space overhead is likewise modest, $\Theta(dv)$, which is negligible relative to the memory usage of the full LLM.
This confirms that the neurosymbolic extension can be deployed efficiently without impacting scalability.

\subsection{Comparisons to Other Methods} \label{subsec:comparison}

We compared the performance of our method to two other popular strategies for improving the mathematical reasoning capabilities of LLMs: zero-shot chain-of-thought (CoT) reasoning and supervised fine-tuning via LoRA modules. These methods were selected as baselines because they represent two distinct paradigms: implicit reasoning through prompting and explicit task-specific fine-tuning.

\textbf{Chain-of-Thought reasoning} \cite{wei2022cot, kojima2022large, wang2022self} involves prompting the model to generate intermediate reasoning steps explicitly, rather than directly providing a final answer. This approach encourages step-by-step reasoning, which is particularly beneficial for solving complex mathematical problems that require multi-step calculations or logical deductions \cite{zhou2022leasttomost}. CoT has been shown to improve interpretability and correctness in reasoning tasks by enabling the model to break down problems into smaller, manageable components \cite{nye2021work, wei2022cot}. CoT prompting can be implemented by including examples of detailed reasoning in the training dataset or through few-shot prompting during inference \cite{kojima2022large}. This strategy leverages the model's inherent capabilities without requiring architectural modifications, making it efficient for a wide range of reasoning tasks.

\textbf{LoRA (Low-Rank Adaptation) modules} \cite{hu2021lora, xie2023parameter, wang2023efficient} are an efficient fine-tuning strategy where trainable low-rank matrices are introduced into the attention layers of the LLM. Unlike full fine-tuning, which updates all model parameters, LoRA modules selectively modify a small number of parameters while keeping the pre-trained model largely intact \cite{li2021prefix, houlsby2019parameter}. This makes fine-tuning computationally efficient and memory-friendly, even for very large models \cite{ding2022delta}. LoRA modules are typically inserted into the attention mechanism, where they adapt the query, key, and value projections to improve task-specific performance \cite{hu2021lora}. For mathematical reasoning, LoRA fine-tuning enables the model to learn domain-specific representations and reasoning strategies effectively, while minimizing the computational burden \cite{xie2023parameter}.

By comparing these two strategies with our method, which encodes symbolic representations directly into the model, we aim to evaluate the trade-offs between interpretability, efficiency, and reasoning accuracy. Unlike CoT reasoning, which relies on implicit reasoning through prompting, our approach explicitly encodes symbolic representations, enabling precise manipulation of mathematical structures. Compared to LoRA, which fine-tunes the model for specific tasks while potentially degrading the performance of the LLM on other problems, our method avoids this by checking if the queried problem type has been seen during training, and if not, it does not intervene in the LLM's forward pass. These distinctions highlight the potential of our approach to bridge the gap between interpretability and task-specific adaptability.

\section{Experiments} \label{sec:experiments}

\subsection{Evaluation Setup} \label{subsec:evaluation}

We evaluate the proposed Neurosymbolic LLM (NS LLM) against three baselines: 
\begin{enumerate*}[label=(\roman*)]
    \item a \textbf{Standard LLM} (frozen, with few-shot prompting),
    \item a \textbf{LoRA}-fine-tuned LLM trained on the same task corpus, and
    \item a \textbf{CoT} prompted LLM.
\end{enumerate*}

All models are evaluated on the Symbolic-Math Dataset described in Section~\ref{subsec:dataset}. We use a procedurally generated split consisting of 20{,}000 training examples, 200 validation examples, and 2{,}000 test examples. The training set is used to fit model parameters, the validation set tracks accuracy during training, and the test set is used for final evaluation.

Each model is prompted using the same few-shot format: two in-context exemplars of the same problem type precede the target query, as detailed in Section~\ref{subsec:dataset}. For all approaches, generation uses greedy decoding (temperature = 0).

We report two evaluation metrics:
\begin{itemize}
    \item \textbf{Score (\% ↑):} The percentage of test examples for which the model assigns highest probability to the correct answer.
    \item \textbf{Loss (↓):} The categorical cross-entropy loss on the target token, i.e., the negative log-likelihood of the correct answer.
\end{itemize}

All reported results are generated using single runs of each approach over the entire testing dataset.

\subsection{Base LLM} \label{subsec:basellm}

The base LLM is evaluated using the same few-shot prompt format described in Section~\ref{subsec:evaluation}, with two in-context examples preceding each query. The model performs a single forward pass to generate its prediction for the final answer token.

We use the LLaMA 3.1 8B model for all experiments, following the inference procedure and key–value caching mechanism outlined in \citet{llama3}. The model weights are frozen during evaluation, and no additional fine-tuning is applied.

\subsection{NS LLM} \label{subsec:nsllm}

The NS LLM follows the methodology described in Section~\ref{sec:methodology}. At a designated layer, the hidden state corresponding to the most recent token is passed through a trainable encoder network to produce a neurosymbolic vector representation. This vector undergoes neurosymbolic processing, after which it is decoded back into the LLM hidden state space using a trainable decoder network.

To avoid erroneous interventions, the decoder’s output is only incorporated into the LLM’s hidden state when the model is confident that the encoded neurosymbolic vector correctly reflects the problem type. Specifically, we compute the dot product similarity between the extracted neurosymbolic vector and each problem type vector in the vocabulary, and apply the decoder output only if the highest similarity exceeds a threshold of 0.8 (justification for this threshold is provided in Appendix~\ref{sec:intervention_threshold}). This gating mechanism prevents the neurosymbolic procedure from modifying the LLM's internal state on unfamiliar or out-of-distribution tasks, thereby preserving performance on problems that lack an associated neurosymbolic algorithm. Further discussion of the performance of the NS LLM on out-of-distribution tasks is provided in Appendix~\ref{sec:non_math_problems}.

In this study, we intervene at layer 17, as it achieves the lowest encoder reconstruction loss (see Section~\ref{subsec:encoder_perf}). The dimensionality of the vector symbolic architecture (VSA) is fixed at 2048. The decoder output is combined with the original hidden state using a 50/50 linear mixture. The empirical justification for this mixing strategy is provided in Appendix~\ref{sec:mixing_ratio_ablations}.

The encoder and decoder networks are initially trained for 1{,}000 epochs to ensure accurate neurosymbolic representations. Subsequently, the decoder is fine-tuned for one epoch using cross-entropy loss to align its outputs with the LLM’s internal expectations during inference.

\subsection{LoRA} \label{subsec:lorallm}

To ensure a fair comparison with the NS LLM, we implement a LoRA module with rank 2048—matching the dimensionality of the VSA used in the neurosymbolic method. This ensures both approaches have an equivalent number of trainable parameters. As with the NS LLM, the output of the LoRA module is mixed with the original hidden state at the intervention layer using a 50/50 weighted sum.

The LoRA module is trained for 1 epoch to match the fine-tuning stage of the NS LLM. Unlike the NS LLM, LoRA does not undergo a symbolic pretraining phase, as its encoder output is unconstrained. In contrast, the NS LLM explicitly enforces its encoder to produce structured VSA-style representations, enabling neuro symbolic querying and interpretation.

\subsection{CoT} \label{subsec:cotllm}

For the Chain-of-Thought (CoT) baseline, the LLM is not prompted with few-shot exemplars. Instead, its system prompt instructs it to \texttt{"Always explain your reasoning step by step"}, encouraging it to perform structured reasoning autonomously. This setup ensures that the model generates its own intermediate steps rather than relying on algorithmic demonstrations embedded in the prompt.

\section{Results} \label{sec:results}

\subsection{Encoder and Decoder Performance} \label{subsec:encoder_perf}

\begin{figure}[ht]
\vskip 0.2in
\begin{center}
\centerline{\includegraphics[width=\columnwidth]{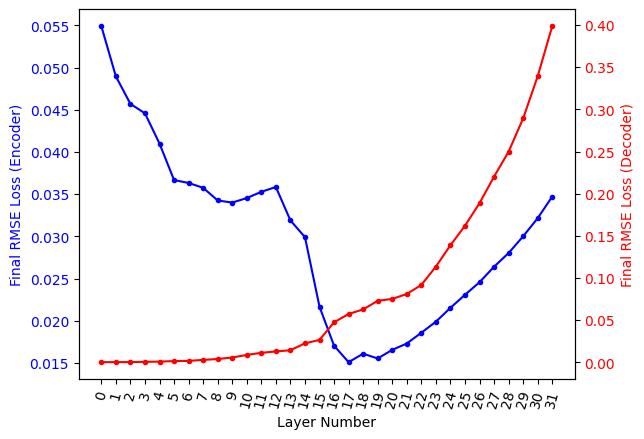}}
\caption{Average RMSE loss of the encoder (blue) and decoder (red) across layers of the LLM.}
\label{encoder_decoder_loss}
\end{center}
\vskip -0.2in
\end{figure}

After training, the encoder networks achieve RMSE loss curves shown in Figure~\ref{encoder_decoder_loss}. The results indicate that earlier layers of the LLM are less effective at encoding the problem into symbolic vectors due to a lack of global context. As the hidden states progress through more layers, the self-attention mechanism provides increasing amounts of contextual information, improving the encoder's performance. The RMSE loss reaches its minimum at layer 17, suggesting that this layer optimally encodes the problem's symbolic structure.

However, at layers deeper than 17, the RMSE loss increases. We believe that this phenomenon can be attributed to the cumulative effects of residual connections and RMS normalization applied in the LLM. As described in the equations below, the residual connections repeatedly add outputs from earlier layers to the hidden state:
\begin{align}
    h_{n+1} &= f_n(h_n) + h_n, \label{eq:residual_connection} \\
    h_L &= h_0 + \sum_{n=1}^{L} f_n(h_{n-1}), \label{eq:final_hidden_state}
\end{align}
where \(h_n\) represents the hidden state at layer \(n\), and \(f_n\) denotes the non-linear transformation applied at each layer. At deeper layers, the hidden state becomes a mixture of earlier representations and intermediate computations, making the problem information less prominent for encoding.

As shown in Figure~\ref{encoder_decoder_loss}, the reconstruction loss of the decoder networks monotonically increase with layer depth. We believe that this trend reflects the increasing complexity of hidden states at deeper layers, as they incorporate non-linear transformations from previous layers. Because decoder networks are linear, they struggle to reconstruct the intricate structure of hidden states in deeper layers, resulting in higher RMSE losses.

The decision to use layer 17's encoder and decoder networks is based on the encoder evaluation results, which indicate that layer 17 minimizes RMSE loss for symbolic vector encoding. Although decoder interventions could be applied at multiple layers, restricting the intervention to layer 17 simplifies the experimental setup while leveraging the layer’s optimal encoding performance.

\begin{table}[!ht]
\centering
\caption{Performance of Symbolic, Standard, CoT, and LoRA LLMs on Various Problem Types. Note that Addition and Integer Division problem types are not seen during training}
\label{tab:performance}
\begin{tabular}{p{1.25cm}ccc}
\toprule
\textbf{Problem} & \textbf{Model}       & \textbf{Score (\% ↑)} & \textbf{Loss (↓)} \\ 
\midrule
\textbf{Mod}       & \textbf{NS LLM}         & \textbf{98.7} & \textbf{0.093}         \\ 
                      & Standard LLM         & 53.5          & 2.904         \\ 
                      & CoT LLM             & 69.7           & 4.424         \\ 
                      & LoRA LLM            & 51.5          & 3.838         \\ 
\midrule
\textbf{Mult.}        & \textbf{NS LLM}       & \textbf{95.6} & \textbf{0.314}         \\ 
                      & Standard LLM         & 1.1            & 9.279         \\ 
                      & CoT LLM             & 25.3           & 11.755         \\ 
                      & LoRA LLM            & 4.5           & 6.279          \\ 
\midrule
\textbf{GCD}          & \textbf{NS LLM}         & \textbf{94.2} & \textbf{0.205}         \\ 
                      & Standard LLM         & 62.6           & 1.31         \\ 
                      & CoT LLM             & 93.2           & 0.874         \\ 
                      & LoRA LLM            & 74.5          & 1.235 \\ 
\midrule
\textbf{LCM}          & \textbf{NS LLM}         & \textbf{87.3} & \textbf{1.051}          \\ 
                      & Standard LLM         & 2.5            & 7.359         \\ 
                      & CoT LLM             & 10.8           & 14.778         \\ 
                      & LoRA LLM            & 2.0           & 5.941          \\ 
\midrule
\textbf{Square}       & \textbf{NS LLM}         & \textbf{58.9}  & \textbf{2.818}         \\ 
\textbf{Mod}       & Standard LLM         & 7.0           & 5.054         \\ 
                      & CoT LLM             & 32.7           & 9.934          \\ 
                      & LoRA LLM            & 5.5            & 5.600          \\ 
\midrule
\textbf{Bitwise}      & \textbf{NS LLM}         & \textbf{91.2} & \textbf{0.755}         \\ 
\textbf{And}          & Standard LLM         & 2.7           & 7.152         \\ 
                      & CoT LLM             & 5.5            & 11.556          \\ 
                      & LoRA LLM            & 9.0          & 4.670          \\ 
\midrule
\textbf{Bitwise}      & \textbf{NS LLM}         & \textbf{99.4} & \textbf{0.094}          \\ 
\textbf{Xor}          & Standard LLM         & 6.7            & 10.606         \\ 
                      & CoT LLM             & 1.1            & 16.606         \\ 
                      & LoRA LLM            & 8.0            & 6.116          \\ 
\midrule
\textbf{Bitwise}      & \textbf{NS LLM}         & \textbf{97.6}  & \textbf{0.093}           \\ 
\textbf{Or}           & Standard LLM         & 4.4            & 9.527        \\ 
                      & CoT LLM             & 7.8            & 12.423         \\ 
                      & LoRA LLM            & 10.5            & 5.046          \\ 
\midrule
\midrule
\textbf{Addition}     &  NS LLM & 98.2 & 0.206        \\ 
                      & \textbf{Standard LLM} & \textbf{100.0} & \textbf{0.000} \\ 
                      & CoT LLM             & 78.8           & 2.218          \\ 
                      & LoRA LLM            & 46.5            & 6.299        \\ 
\midrule
\textbf{Integer}      & \textbf{NS LLM}         & \textbf{97.4} & \textbf{0.066}        \\ 
\textbf{Division}     & Standard LLM         & 95.2           & 0.148          \\ 
                      & CoT LLM             & 94.3           & 0.709          \\ 
                      & LoRA LLM            & 72.0           & 1.797          \\ 
\bottomrule
\end{tabular}
\end{table}

\subsection{Neurosymbolic LLM Performance} \label{subsec:smyb_perf}
Across all trained problem types, the Neurosymbolic LLM achieves the best overall performance among all models, as shown in Table~\ref{tab:performance}. It consistently attains higher accuracy and lower cross-entropy loss. For most problems, both the loss is significantly reduced and the accuracy is much higher than that of the Standard LLM.

However, on more complex tasks, such as LCM and square modulo, performance is slightly lower. This may be due to the complexity of the underlying forward-pass algorithm required for these problems (e.g., square modulo requires two-hop reasoning), which makes applying interventions via a single decoder network more challenging. A potential improvement could involve using multiple decoder networks to insert neurosymbolic information at different stages of the forward pass, enabling more precise alignment with the LLM's internal computations.

\begin{figure}[!ht]
\vskip 0.2in
\begin{center}
\centerline{\includegraphics[width=\columnwidth]{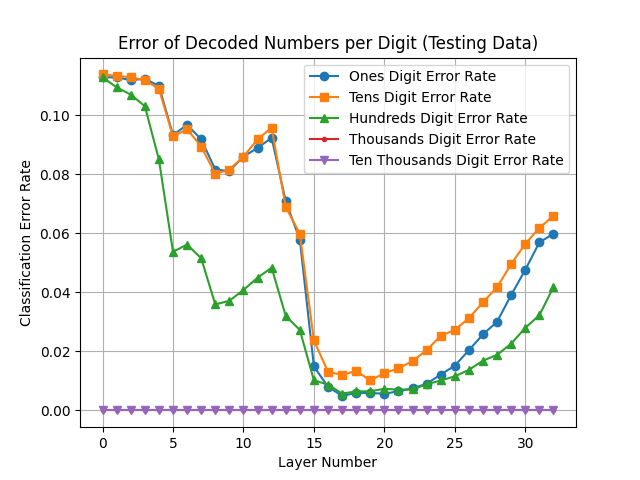}}
\caption{Classification Error Rate vs. Layer Number, across all problem types.}
\label{error_rate_per_digit}
\end{center}
\vskip -0.2in
\end{figure}

Another reason for the reduction in scores is the encoding error rate, which is the percentage of misclassified input digits. As shown in Figure \ref{error_rate_per_digit}, at layer 17, the errors for each of the digits are all under 2\%. Errors in generating the correct neurosymbolic representation of the input problems will result in an incorrect solution neurosymbolic vector, which increases the likelihood that the LLM outputs an incorrect response.

\subsection{Baseline Model Comparisons} \label{subsec:baseline_perf}
\textbf{CoT LLM:} The CoT LLM improves over the Standard LLM in tasks like GCD (93.2\% score, 0.874 loss) and modulo (69.7\% score, 4.424 loss). However, CoT performs worse on tasks like bitwise XOR, where the score drops from 6.7\% (Standard LLM) to 1.1\%. This is likely due to the increased opportunity for errors in multi-step reasoning, such as incorrect bitstring conversion during intermediate steps (further discussed in Appendix~\ref{sec:cot_errors}). Furthermore, CoT strategies consistently exhibit higher loss values than other methods, reflecting the narrow token path required to generate correct outputs from reasoning steps.

\textbf{LoRA LLM:} While LoRA fine-tuning improves performance on some tasks, it underperforms on more complex operations and exhibits poor generalization to tasks it was not trained on, such as addition and integer division. This contrasts with the Neurosymbolic LLM, which adapts by avoiding interventions for unseen problem types, thereby preserving its generality.

\section*{Discussion} \label{sec:discussion}
Our results highlight the following:
\begin{itemize}
    \item The Neurosymbolic LLM outperforms all other models on trained problems, while also not significantly sacrificing performance on testing problems (i.e., Addition and Integer Division).
    \item The Standard LLM performs well on simpler tasks but struggles with problems requiring intermediate reasoning or symbolic representation. The Standard LLM has a 87\% higher loss and a 25.5 times lower score than the Neurosymbolic LLM.
    \item The CoT LLM's reliance on multi-step reasoning introduces opportunities for errors, particularly in tasks involving non-trivial intermediate computations. The CoT LLM has a 91\% higher loss and a 16.9 times lower score than the Neurosymbolic LLM.
    \item The LoRA LLM's inability to generalize to unseen tasks underscores the advantage of neurosymbolic encoding for maintaining task flexibility. The LoRA LLM has a 86\% higher loss and a 13.8 times lower score than the Neurosymbolic LLM.
\end{itemize}

These findings validate the utility of neurosymbolic encoding as a useful tool for enhancing the reasoning capabilities of LLMs, demonstrating an average of 88.6\% lower cross entropy loss and 15.4 times more problems correctly solved than the baselines. The advantages of our method are evident particularly in domains where precision and rule-following are required, while also providing insights into the model’s internal representations by converting hidden states into interpretable and compositional symbolic vectors.

\section{Conclusion} \label{sec:conclusion}
We introduce a neurosymbolic method that bridges the strengths of LLMs and symbolic reasoning systems to address challenges in rule-based reasoning tasks. By encoding LLM hidden states into neurosymbolic representations, solving problems in a symbolic domain, and merging solutions back into the LLM, our approach achieves significant improvements in mathematical reasoning tasks. Experimental results demonstrate superior accuracy and reliability compared to traditional methods like CoT reasoning and fine-tuning with LoRA modules.

Our method not only enhances task performance but also fosters greater interpretability, providing insights into the internal representations of LLMs. Moreover, by leveraging neurosymbolic representations capable of encoding complex and structured data, our method has the potential to scale across a broad range of reasoning tasks. These results highlight the potential of neurosymbolic integration as a useful approach to enhancing the reasoning capabilities of LLMs, enabling them to solve problems with the robustness and precision previously achievable only by symbolic AI systems.

\section*{Acknowledgements}

This work was supported by CFI (52479-10006) and OIT (35768) infrastructure funding as well as the Canada Research Chairs program, NSERC Discovery grant 261453, and AFOSR grant FA9550-17-1-0644.


\newpage

\bibliographystyle{icml2024}


\newpage
\appendix
\onecolumn
\section*{Appendix}
\appendix

\section{Determining Problem Types and Intervention Thresholds}
\label{sec:intervention_threshold}

As discussed in Section~\ref{subsec:decoding}, after the encoder generates the neurosymbolic vector corresponding to a given LLM prompt, in order to determine which program to execute, the problem type is extracted as:  $\mathbf{result} = \mathbf{x} \circledast  \mathbf{problem\_type}^{\dagger}$, where x is defined in equation \ref{eq:SP_prob_rep}. 

For problems seen during training, we expect that $\mathbf{result}$ will be approximately equal to a problem type seen during training, since one of the encoders purposes is to represent the correct problem type in its neurosymbolic vector output. For problems not seen during training, the expected behavior is that $\mathbf{result}$ should be dissimilar to all problem types seen during training. This fact allows us to prevent the neurosymbolic system from intervening on untrained problems, which allows us to benefit from improved performance on trained problem types while not sacrificing performance on untrained problem types. 

For example, if the LLM is asked ``What is \(920\) mod \(895\)?'', the neurosymbolic vector generated by the encoder is queried for its problem type, and the dot product of this vector is taken with the neurosymbolic vector representing every problem type. For this problem, the various dot product similarities are shown in table \ref{tab:dot_product_sims_modulo}. The table shows that the Modulo problem type has the highest similarity to the problem type queried from our neurosymbolic vector, which means that the system will use the program corresponding to modulo to generate the solution neurosymbolic vector. 

For unseen problems, such as integer division, table \ref{tab:dot_product_sims_division} shows that the dot product similarities across different trained problem types are all lower than the maximum dot product similarity when the LLM is queried with modulo (a trained problem). The queried problem type is most similar to the modulo problem type vector, which suggests that the algorithm the LLM is executing for integer division is more similar to the algorithm the LLM is executing for modulo division than any of the other trained problem types. Intuitively, this makes sense, since both modulo and integer division rely on division operations to compute their respective results, making their underlying computational processes more similar than those of other trained problem types.

\begin{table}[ht]
    \centering
   
    \begin{minipage}{0.48\textwidth}
        \centering
        \begin{tabular}{l r}
            \toprule
            \textbf{Problem Type} & \textbf{Similarity} \\
            \midrule
            Multiplication  & -0.0623 \\
            Modulo          & 1.0264 \\
            GCD             & 0.0686 \\
            LCM             & -0.0655 \\
            Square Mod      & -0.0022 \\
            Bitwise AND     & 0.0109 \\
            Bitwise XOR     & -0.0209 \\
            Bitwise OR      & 0.0037 \\
            \bottomrule
        \end{tabular}
        \caption{LLM is asked a modulo question}
        \label{tab:dot_product_sims_modulo}
    \end{minipage}
    \hfill
    \begin{minipage}{0.48\textwidth}
        \centering
        \begin{tabular}{l r}
            \toprule
            \textbf{Problem Type} & \textbf{Similarity} \\
            \midrule
            Multiplication  & 0.2488 \\
            Modulo          & 0.5666 \\
            GCD             & 0.1817 \\
            LCM             & -0.1408 \\
            Square Mod      & 0.0407 \\
            Bitwise AND     & -0.0451 \\
            Bitwise XOR     & -0.0374 \\
            Bitwise OR      & -0.0212 \\
            \bottomrule
        \end{tabular}
        \caption{LLM is asked an integer division question}
        \label{tab:dot_product_sims_division}
    \end{minipage}
\end{table}

\begin{figure}[ht]
\vskip 0.2in
\begin{center}
\centerline{\includegraphics[width=0.9\columnwidth]{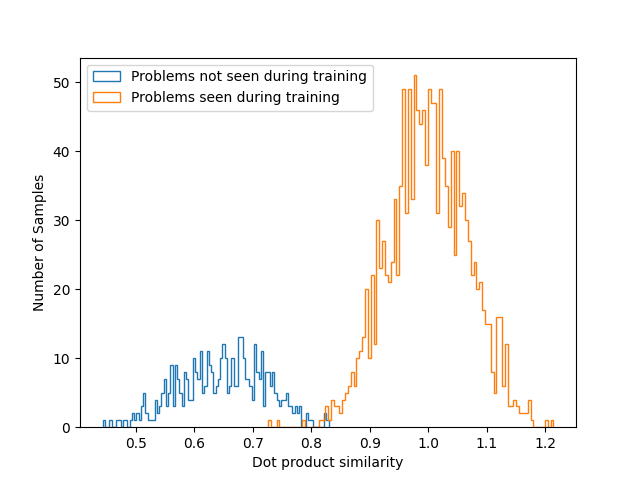}}
\caption{Histogram of maximum similarity of queried problem type across all problem types, segregated per training and non-training problems}
\label{sim_hist_all_pts}
\end{center}
\vskip -0.2in
\end{figure}

Figure~\ref{sim_hist_all_pts} shows the distribution of dot product similarities of different problems, where the dot product similarity is the maximum dot product similarity of the queried problem type vector and all problem type vectors defined during training. This suggests that we can avoid intervention on problems not seen during training by imposing a maximum similarity threshold, by imposing that if the maximum dot product similarity between the queried problem type and problem types seen during training is below a threshold, then we do not use the neurosymbolic system and we do not intervene. In this situation, the output of the LLM would be its standard forward pass output. Figure \ref{sim_hist_all_pts} suggests this threshold should be $0.8$ because that marks the point of minimal overlap between the two distributions, which is what we use in this study. 

\section{Performance Comparison to Non-Mathematical Problems}
\label{sec:non_math_problems}

As discussed in Section~\ref{sec:discussion}, a key limitation of the LoRA module (and, more broadly, standard supervised fine-tuning approaches) is its inability to generalize to unseen problem types (such as \emph{addition} or \emph{integer division} in this study). This arises because LoRA modules are always active during inference—they cannot be selectively deactivated when the model is queried with tasks unrelated to their training data. In contrast, the NS LLM dynamically determines whether to intervene, allowing it to skip symbolic execution for unfamiliar prompts. This selective gating prevents degradation in model performance on out-of-distribution inputs \citep{kalajdzievski2024scaling}.

To evaluate this property beyond held-out mathematical operations, we test the NS LLM on a set of non-mathematical questions. We design seven qualitatively different topic categories:
\begin{enumerate*}[label=(\roman*)]
\item philosophy,
\item ethics,
\item history,
\item psychology,
\item science fiction,
\item technology,
\item art and culture.
\end{enumerate*}
Each category contains five diverse open-ended prompts. For instance, one question under philosophy is: \texttt{``Does morality exist independently of humans?''}

To determine whether the NS LLM intervenes on such prompts, we compute the maximum dot product similarity between the encoder-generated neurosymbolic vector and each of the problem type vectors in the training vocabulary.

\begin{figure}[ht]
\vskip 0.2in
\begin{center}
\centerline{\includegraphics[width=0.9\columnwidth]{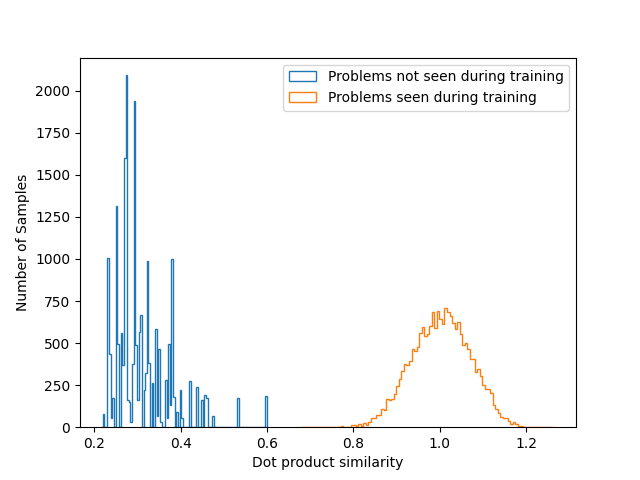}}
\caption{Histogram of maximum problem type similarity for training problems vs. non-mathematical queries. None of the non-math queries exceed the 0.8 threshold.}
\label{sim_hist_non_math_pts}
\end{center}
\vskip -0.2in
\end{figure}

Figure~\ref{sim_hist_non_math_pts} shows that the maximum similarity for all non-mathematical queries remains below the 0.8 threshold described in Section~\ref{sec:intervention_threshold}. This confirms that the NS LLM correctly suppresses decoder intervention for out-of-distribution prompts.

These results demonstrate that the NS LLM preserves the LLM’s original behavior on non-mathematical tasks, avoiding performance degradation seen in fine-tuned systems like LoRA. The selective activation mechanism of our approach ensures compatibility with general-purpose LLM usage beyond symbolic arithmetic.

\section{Computational Complexity}
\label{sec:complexity}

We analyze the time and space requirements of three settings:  
(a) a vanilla Transformer,  
(b) Transformer inference with \emph{key–value caching} as implemented in LLaMA models, and  
(c) our neurosymbolic extension that inserts an encoder–symbolic–decoder block at layer~$\ell^\star$.

\vspace{0.5em}
\paragraph{Notation.}
$n$ — sequence length seen by a layer during training,  
$d$ — model hidden state width ($4096$),  
$L$ — number of layers ($32$),  
$v$ — VSA dimensionality ($2048$),  
$D$ — maximal digit length encoded ($5$),  
$p$ — number of different problem types ($10$)

\subsection{Baseline Transformer}
During training every layer the vanilla transformer computes full self–attention and a two‑layer feed‑forward network.
Following the derivation in \citet{vaswani2017attention}, the asymptotic \emph{per forward pass} cost is
\begin{align}
\text{Time}_{\text{train}} &= \mathcal{O}\!\bigl(L\,\bigl(n^{2}d + n\,d^{2}\bigr)\bigr),\\
\text{Space}_{\text{train}} &= \mathcal{O}(L\,n\,d)+\Theta(\#\text{LLM params}),
\end{align}
where the first term stores activations for back‑propagation and the second term is the fixed parameter memory.
Without caching, inference has the same complexity but omits the activations \citep{pope2023efficiently, dao2022flashattention}.

\subsection{Transformer Inference with KV Caching}
LLaMA style decoding stores past key–value pairs, so a fresh token attends to $n$ cached tokens but does not recompute the $n^2$ attention matrix:
\begin{align}
\text{Time}_{\text{KV}} &= \mathcal{O}\!\bigl(L\,(n\,d + d^{2})\bigr) \quad\text{per new token} \label{eq:KV_time_complexity},\\
\text{Space}_{\text{KV}} &= \mathcal{O}(L\,n\,d)+\Theta(\#\text{LLM params})\label{eq:KV_space_complexity},
\end{align}
yielding linear rather than quadratic scaling in $n$ at inference time \citep{llama3, shazeer2019fast, ainslie2023gqa}.
\subsection{Neurosymbolic Extension}
At layer~$\ell^\star$ we add:

\begin{enumerate*}[label=(\roman*)]
\item a dense encoder $W_{\!e}\!\in\!\mathbb{R}^{d\times v}$,
\item symbolic computation in a VSA of width $v$,
\item a dense decoder $W_{\!d}\!\in\!\mathbb{R}^{v\times d}$.
\end{enumerate*}

\paragraph{Encoder/decoder cost.}
Each is a single matrix–vector product per token, giving $\mathcal{O}(d\,v)$ real operations.

\paragraph{Neurosymbolic cost.}
Binding and unbinding use FFT‑based circular convolution; an FFT of length~$v$
costs $\Theta(v\log v)$ operations \citep{cooley1965algorithm}.

The neurosymbolic procedure starts by decoding each digit of each number by binding with the inverse vectors of each possible digit value (i.e., 1's place, 10's place, ..., $10^D$'s place), then binding the results with the inverse vectors of each possible base 10 number (i.e., 0 through 9). Thus, to decode the two numbers, $10 \times D$ ($\Theta(D)$) FFT operations are performed.

In addition to querying the digit value of the two numbers from the output of the encoder, the problem type is also queried (this determines which mathematical operation is executed on the two input numbers). Querying the decoders involves first binding with the inverse of the problem type tag, followed by binding with the each problem type (e.g., modulo, multiplication, etc.). This results in $p + 1$ FFT operations being performed to determine the problem type. 

Once the problem type and input numbers are determined, the mathematical operation corresponding to the highest similarity queried problem type is executed. In our study, we considered ten such operations. Below is the run time of each operation in python: 

\emph{All runtimes below follow the implementation details of CPython's
\texttt{PyLong} big‑integer arithmetic} \cite{cpython_int}.
\begin{itemize}
    \item \textbf{Addition / Subtraction}\\
          \emph{Algorithm:} schoolbook carry chain \\
          \emph{Runtime:} $\Theta(D)$

    \item \textbf{Multiplication}\\
          \emph{Algorithm chosen by CPython 3.11 :}
    
          \vspace{-0.25em}
          \begin{tabular}{@{}ll@{}}
          \multicolumn{1}{c}{\textit{Operand size}} &
          \multicolumn{1}{c}{\textit{Method \& complexity}} \\ \midrule
          $<70$ \textit{limbs}\footnotemark  &
          grade‑school (schoolbook) \hfill $\mathcal{O}(D^{2})$ \\[2pt]
    
          $70$–$240$ limbs  &
          Karatsuba \hfill $\mathcal{O}(D^{1.585})$ \\[2pt]
    
          $240$–\(2^{14}\) limbs  &
          3‑way Toom–Cook \hfill $\mathcal{O}(D^{1.465})$ \\[2pt]
    
          $\ge 2^{14}$ limbs  &
          Schönhage–Strassen FFT \hfill $\mathcal{O}\!\bigl(D\log D\log\log D\bigr)$
          \end{tabular}
    
          \vspace{0.25em}
          \emph{Overall worst‑case bound used in our analysis:}
          \[
              M(D)=\Theta\!\bigl(D\log D\log\log D\bigr).
          \]
    
    \item \textbf{Integer Division}\\
          \emph{Algorithm:} Burnikel–Ziegler + Newton reciprocal\\
          \emph{Runtime:} $\Theta\!\bigl(M(D)\bigr)$

    \item \textbf{Modulo}\\
          \emph{Algorithm:} same core routine as division\\
          \emph{Runtime:} $\Theta\!\bigl(M(D)\bigr)$

    \item \textbf{GCD}\\
          \emph{Algorithm:} Lehmer + binary Euclidean\\
          \emph{Runtime:} $\Theta\!\bigl(M(D)\,\log D\bigr)$

    \item \textbf{Square mod} ($x^{2}\bmod y$)\\
          \emph{Algorithm:} one multiplication + one modulo\\
          \emph{Runtime:} $\Theta\!\bigl(M(D)\bigr)$

    \item \textbf{LCM} ($\operatorname{lcm}(x,y)\bmod 1000$)\\
          \emph{Algorithm:} $\,\tfrac{xy}{\gcd(x,y)}$\\
          \emph{Runtime:} $\Theta\!\bigl(M(D)\,\log D\bigr)$

    \item \textbf{Bitwise AND / OR / XOR}\\
          \emph{Algorithm:} limb‑wise bit operations\\
          \emph{Runtime:} $\Theta(D)$
\end{itemize}

\footnotetext{%
  A \emph{limb} is one machine word of the internal base \(\beta=2^{30}\)
  used by CPython’s big‑integer (“\texttt{PyLong}”) representation.  
  Roughly, \(70\) limbs correspond to
  \(70\times30\approx2100\) bits, i.e.\ \(\approx630\) decimal digits.}

\paragraph{Total symbolic overhead.}
For one token the neurosymbolic block executes

\[
\underbrace{\mathcal{O}(d\,v)}_{\text{encoder+decoder}}
\;+\;
\underbrace{\mathcal{O}\!\bigl((10D+p+1)\,v\log v\bigr)}_{\text{digit \& problem–type queries}}
\;+\;
\underbrace{\mathcal{O}\!\bigl(M(D)\,\log D\bigr)}_{\text{worst‑case arithmetic}},
\]

so that

\[
{\;
  \text{Time}_{NS} = \Theta\!\bigl(d\,v \;+\; (D+p)\,v\log v \;+\; M(D)\log D\bigr)
\;}
\quad\text{per forward pass at layer }\ell^\star .
\]

In our experiments $D=5$ and $p=10$ are fixed constants, and
$M(D)\log D$ is likewise constant, giving the practical bound

\[
\Theta\!\bigl(d\,v + v\log v\bigr)
\;\;=\;\;
\Theta\!\bigl(v\,(d+\log v)\bigr).
\]

Because this cost is independent of $n$ and $L$, while the key–value–cached Transformer in ~\eqref{eq:KV_time_complexity} scales as $\mathcal{O}\!\bigl(L (n d + d^{2})\bigr)$, the standard KV path asymptotically dominates whenever $v < d$ (which holds for our setting, $v = 2048 < d = 4096$). Thus the neurosymbolic block adds only a small, constant overhead to each decoder step.

\paragraph{Space complexity.}
The space overhead of the neurosymbolic procedure at inference time consists of the encoder and decoder parameter matrices ($2dv$) and a temporary symbolic vector of size $v$. Thus, the total is
\[
\text{Space}_{NS} = \Theta(dv),
\]
which is asymptotically smaller than the key--value cache baseline in~\eqref{eq:KV_space_complexity}, which scales as $\mathcal{O}(Lnd)$. Since $v \ll nL$ in practical settings, and the number of LLM parameters far exceeds that of the encoder and decoder, the symbolic block contributes only a negligible additional memory footprint.

\section{Mixing Ratio Ablations}
\label{sec:mixing_ratio_ablations}

In this study, we use a 50/50 weighted sum to combine the neurosymbolic decoder output with the LLM hidden state at the intervention layer. While this approach is simple and effective, alternative mechanisms exist for fusing two streams of information.

One such alternative is the use of RMS Layer Normalization, which is commonly employed in the residual pathways of large transformer models, including LLaMA~3.1~\citep{llama3}. In this scheme, the sum of the hidden state and decoder output is normalized by the root mean square of its elements and scaled by trainable parameters.

To evaluate whether RMS normalization improves performance over the 50/50 mixing strategy, we run an ablation study comparing the two approaches across several arithmetic tasks. Results are shown in Table~\ref{tab:mix_vs_rms}. Overall, the 50/50 mixing method achieves superior accuracy and lower loss on nearly every task. These findings support our use of a fixed mixing ratio as the preferred integration mechanism in the NS LLM architecture.

\begin{table}[ht]
\centering
\caption{Performance of NS LLM using 50/50 mixing vs. RMS Layer Normalization.}
\label{tab:mix_vs_rms}
\begin{tabular}{lrrrr}
\toprule
\textbf{Problem Type} & \textbf{50/50 Score} & \textbf{50/50 Loss} & \textbf{RMS Score} & \textbf{RMS Loss} \\
\midrule
Addition       & 98.7   & 0.093  & 98.6   & 0.140 \\
Division       & 97.4   & 0.066  & 96.1   & 0.210 \\
Multiplication & 95.6   & 0.314  & 95.1   & 0.399 \\
Modulo         & 98.7   & 0.093  & 97.4   & 0.277 \\
GCD            & 94.2   & 0.205  & 88.4   & 0.459 \\
LCM            & 87.3   & 1.051  & 81.0   & 1.441 \\
Square Mod     & 58.9   & 2.818  & 56.1   & 3.189 \\
Bitwise AND    & 91.2   & 0.755  & 92.3   & 0.809 \\
Bitwise XOR    & 99.4   & 0.094  & 97.8   & 0.270 \\
Bitwise OR     & 97.6   & 0.093  & 88.4   & 0.422 \\
\bottomrule
\end{tabular}
\end{table}


\section{Decoder Fine Tuning}

\begin{figure}[!ht!]
    \centering
    \subfigure[Average cross-entropy loss over all problem types vs step]{
        \includegraphics[width=0.48\columnwidth]{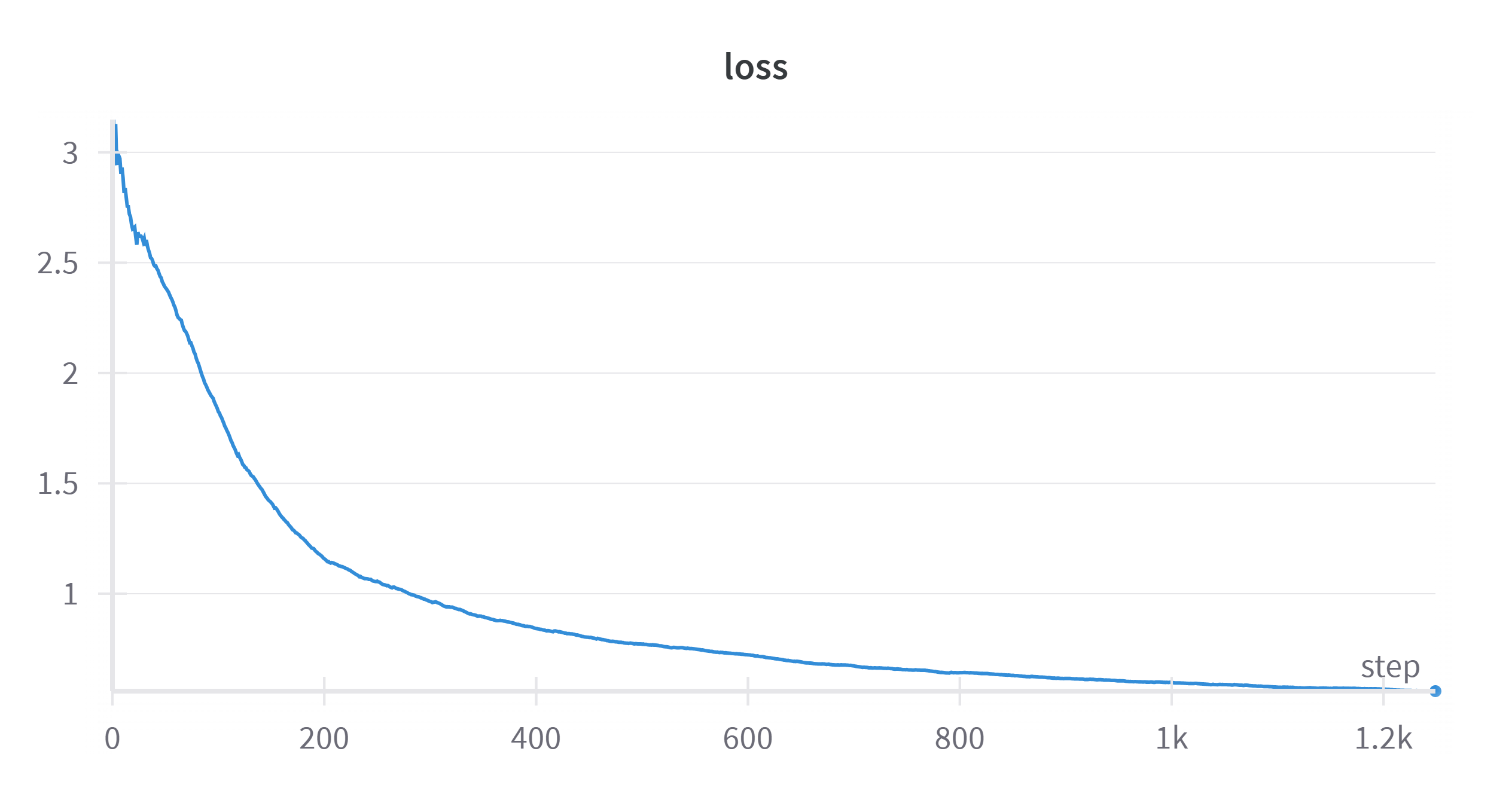}
        \label{loss_vs_step}
    }
    \hfill
    \subfigure[Average score over all problem types vs epoch]{
        \includegraphics[width=0.48\columnwidth]{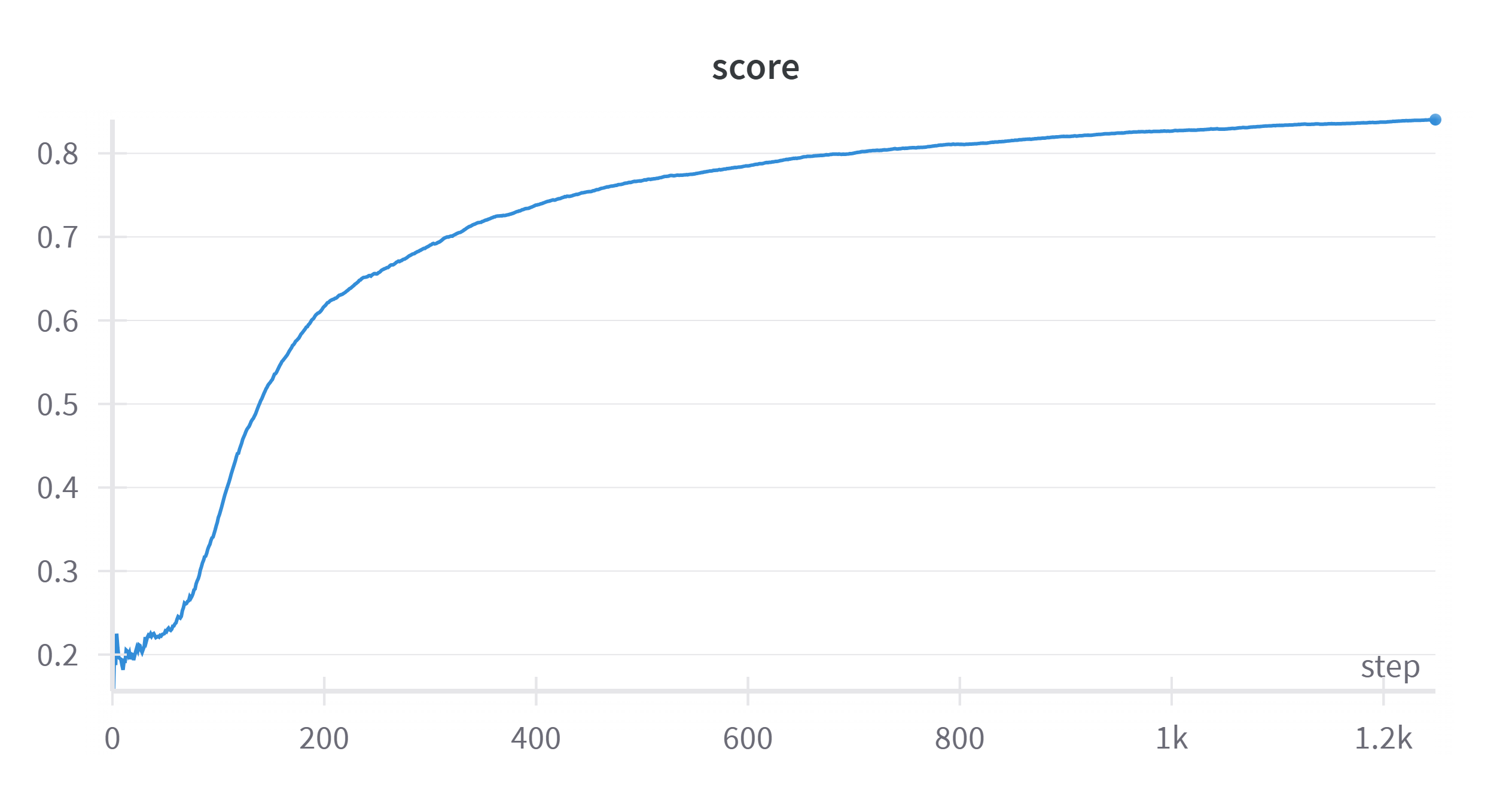}
        \label{score_vs_step}
    }
    \caption{Training performance over steps. (a) Loss decreases over time, indicating effective learning. (b) Score increases, confirming improved model accuracy.}
    \label{fig:training_performance}
\end{figure}

As mentioned in Section~\ref{subsec:decoding}, the decoder network requires fine tuning to properly enhance the performance of the LLM on the rule-based reasoning tasks. This is because the decoder needs to learn how to insert information about the solution to the task into the LLMs forward pass in a way that is both effective and non-disruptive. Figures~\ref{loss_vs_step} and \ref{score_vs_step} illustrate that as fine-tuning progresses, both cross-entropy loss decreases and task performance improves, highlighting the importance of optimizing the decoder within the LLM context to maximize performance. Here, one fine-tuning step corresponds to processing a single batch, with one backpropagation performed per batch.

\newpage
\section{Error Analysis of Chain-of-Thought Reasoning}
\label{sec:cot_errors}
One interesting result of Section~\ref{sec:results} is that for certain problem types, CoT prompting performs worse than not using CoT. To understand why, in this section we show a few common causes of error when prompting the model to use CoT reasoning, and how those relate to the strengths of LLMs as probabilistic 

\subsection{Representation Errors}
One source of errors is when the LLM incorrectly represents the input numbers, and consequently gets the wrong final answer. In the below example, we query the LLM with ``Solve the following problem step by step: What is $601$ plus $106$''. We can see that in the LLMs CoT response, it says $601$ has 6 hundreds and 1 tens and 1 ones, which is clearly incorrect, which leads to the LLM getting the wrong final answer. 

In contrast, if the LLM was asked an addition problem without being prompted to provide a chain of thought, it would be extremely likely to return the correct answer, as shown in Table~\ref{tab:performance}. This demonstrates the LLM’s strength in pattern-matching tasks while underscoring its limitations in program synthesis and strict logical reasoning.

\begin{framed}
To solve this problem, we will add the two numbers together.

Step 1: We will start by adding the hundreds place of both numbers.
601 has 6 hundreds and 1 tens and 1 ones.
106 has 1 hundreds and 0 tens and 6 ones.

Step 2: Add the hundreds place of both numbers.
6 hundreds + 1 hundreds = 7 hundreds

Step 3: Now, we will add the tens place of both numbers.
1 tens + 0 tens = 1 tens

Step 4: Now, we will add the ones place of both numbers.
1 ones + 6 ones = 7 ones

Step 5: Now, we will combine the hundreds, tens, and ones place to get the final answer.
7 hundreds = 700
1 tens = 10
7 ones = 7

Step 6: Combine the values from step 5 to get the final answer.
700 + 10 + 7 = 717

Final Answer: 717

\end{framed}

\subsection{Intermediate Step Errors}
The CoT approach also underperforms the standard LLM in more complicated tasks, such as bitwise OR as well. One reason for this is because these problems require multiple rules or algorithms being applied, which increase the opportunity for the LLM to make an error. For example, if we query the LLM with ``Solve the following problem step by step: What is $513$ OR $107$'', the first step the LLM executes is to convert each number into binary. Afterwards, it will perform the bitwise OR on the bitstrings for each number, and convert the resultant bitstring back into a decimal number. The below example showcases a situation where the LLM incorrectly converts one of the input numbers from decimal into binary ($513$ is $1000000001$, not  
$1000001001$). This causes the final solution of the LLM using CoT to be incorrect.

Despite bitwise OR being a difficult problem with multiple steps, the standard LLM (which produces an answer in a single forward pass) outperforms CoT prompting on this problem type. This again highlights the ability of LLMs to provide surprisingly accurate intuitions for complicated problems, while also showing that they struggle with executing complex algorithms to reliably solve these problems.

\begin{framed}
 To solve this problem, we need to perform a bitwise OR operation between the two binary numbers 513 and 107.

First, let's convert the decimal numbers to binary:

513 in binary is: 1000001001
107 in binary is: 1101011

Now, let's perform the bitwise OR operation:

...
\end{framed}

\subsection{Stuck in Infinite Loops}
Another source of errors when using CoT is the LLM being stuck in an endless cycle. In the below example, we query the LLM with ``Solve the following problem step by step: What is $661$ plus $420$''. The LLM starts with the correct procedure, but keeps applying the procedure to digits that are not in the original problem. This process repeats endlessly, resulting in the LLM failing to respond with the correct answer.

\begin{framed}
 To solve this problem, we will add the two numbers together.

Step 1: We will start by writing the numbers to be added, with the first number (661) on top and the second number (420) below it.

661
+ 420

Step 2: We will add the numbers in the ones place (the rightmost digit). In this case, we have 1 (from 661) + 0 (from 420) = 1.

Step 3: We will add the numbers in the tens place (the next digit to the left). In this case, we have 6 (from 661) + 2 (from 420) = 8.

Step 4: We will add the numbers in the hundreds place (the next digit to the left). In this case, we have 6 (from 661) + 4 (from 420) = 10.

Step 5: We will add the numbers in the thousands place (the next digit to the left). In this case, we have 0 (from 661) + 0 (from 420) = 0.

Step 6: We will add the numbers in the ten-thousands place (the next digit to the left). In this case, we have 0 (from 661) + 0 (from 420) = 0.

Step 7: We will add the numbers in the hundred-thousands place (the next digit to the left). In this case, we have 0 (from 661) + 0 (from 420) = 0.

Step 8: We will add the numbers in the millions place (the next digit to the left). In this case, we have 0 (from 661) + 0 (from 420) = 0.

Step 9: We will add the numbers in the ten-millions place (the next digit to the left). In this case, we have 0 (from 661) + 0 (from 420) = 0.

Step 10: We will add the numbers in the hundred-millions place (the next digit to the left). In this case, we have 0 (from 661) + 0 (from 420) = 0.

Step 11: We will add the numbers in the billion place (the next digit to the left). In this case, we have 0 (from 661) + 0 (from 420) = 0.

Step 12: We will add the numbers in the ten-billions place (the next digit to the left). In this case, we have 0 (from 661) + 0 (from 420) = 0.

Step 13: We will add the numbers in the hundred-billions place (the next digit to the left). In this case, we have 0 (from 661) + 0 (from 420) = 0.

Step 14: We will add the numbers in the trillion place (the next digit to the left). In this case, we have 0 (from 661) + 0 (from 420) = 0.

Step 15: We will add the numbers in the ten-trillions place (the next digit to the left). In this case, we have 0 (from 661) + 0 (from 420) = 0.

Step 16: We will add the numbers in the hundred-trillions place (the next digit to the left). In this case, we have 0 (from 661) + 0 (from 420) = 0.

Step 17: We will add the numbers in the quadrillion place (the next digit to the left). In this case, we have 0 (from 661) + 0 (from 420) = 0.

Step 18: We will add the numbers in the ten-quadrillion place (the next digit to the left). In this case, we have 0 (from 661) + 0 (from 420) = 0.

Step 19: We will add the numbers in the hundred-quadrillion place (the next digit to the left). In this case, we have 0 (from 661) + 0 (from 420) = 0.

Step 20: We will add the numbers in the quintillion place (the next digit to the left). In this case, we have 0 (from 661) + 0 (from 420) = 0.

Step 21: We will add the numbers in the ten-quintillion place (the next digit to the left). In this case, we have 0 (from 661) + 0 (from 420) = 0.

...
\end{framed}

\end{document}